\documentclass[conference]{IEEEtran}
\usepackage{lingmacros}
\usepackage{tree-dvips}
\usepackage{graphicx}
\usepackage{multirow}
\usepackage{subcaption}
\usepackage{pbox}
\usepackage{array}
\usepackage{float}
\usepackage{changepage}
\usepackage{xcolor,colortbl}
\usepackage[cmex10]{amsmath}
\usepackage{rotating}
\usepackage{url}
\usepackage[utf8]{inputenc}

\ifCLASSINFOpdf
\else
\fi
\begin{document}
%
\title{Recognizing Descriptive Wikipedia Categories for Historical Figures}


%
\author{\IEEEauthorblockN{Yanqing Chen}
\IEEEauthorblockA{Computer Science Department\\
Stony Brook University\\
Stony Brook, New York, 11790\\
Email: cyanqing@cs.stonybrook.edu}
\and
\IEEEauthorblockN{Steven Skiena}
\IEEEauthorblockA{Computer Science Department\\
Stony Brook University\\
Stony Brook, New York, 11790\\
Email: skiena@cs.stonybrook.edu}}



%


\maketitle

\begin{abstract}
Wikipedia is a useful knowledge source that benefits many applications in language processing and knowledge representation. An important feature of Wikipedia is that of categories. Wikipedia pages are assigned different categories according to their contents as human-annotated labels which can be used in information retrieval, ad hoc search improvements, entity ranking and tag recommendations. However, important pages are usually assigned too many categories, which makes it difficult to recognize the most important ones that give the best descriptions.

In this paper, we propose an approach to recognize the most descriptive Wikipedia categories. We observe that historical figures in a precise category presumably are mutually similar and such categorical coherence could be evaluated via texts or Wikipedia links of corresponding members in the category. We rank descriptive level of Wikipedia categories according to their coherence and our ranking yield an overall agreement of 88.27\% compared with human wisdom.

\end{abstract}


%
\IEEEpeerreviewmaketitle

\section{Introduction}




Wikipedia is a useful knowledge source that benefits many applications in language processing and knowledge representation. An important feature of Wikipedia is that of categories. Wikipedia pages are assigned different categories according to their contents as human-annotated labels which can be used in information retrieval, ad hoc search improvements, entity ranking and tag recommendations. 

\begin{figure}[!htb]
\centering
\includegraphics[width=0.5\textwidth]{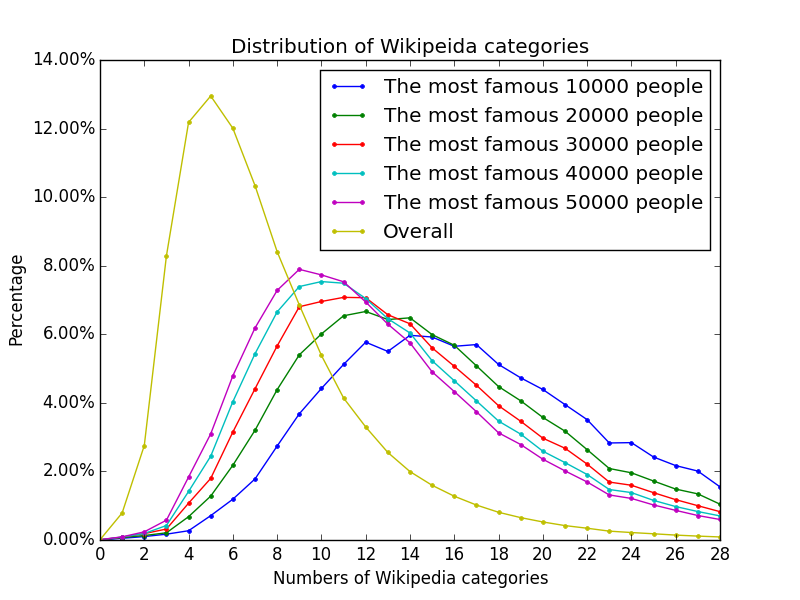}
\caption{Distribution of Wikipedia categories on pages of people. We use ``ranking of significance" described in \cite{skiena2013s} to decide the most famous people on Wikipedia. Overall historical figures have about 8 categories while the most famous 10,000 people on average have more than 20.}
\label{wikicat}
\end{figure}

However, important pages are usually assigned too many categories. Figure \ref{wikicat} shows the distribution of categories for historical figures on Wikipedia.

On the other hand, most of these categories are not descriptive enough. For instance, Barack Obama is listed in almost fifty different Wikipedia categories, including `` 20th-century American writers", ``1961 births", ``American Nobel laureates", `` Grammy Award winners" and ``Harvard Law School alumni". These categories are undoubtedly correct but are sometimes trivial. We believe that ``Presidents of the United States" most accurately captures his historical/cultural significance.

One approach might be based on keywords, e.g. recognizing that ``President" is more important distinction than other titles. However, this appears quite challenging. Being President of a small organization is generally less impressive. Example of Jonathan Edwards shows that even ``President of Princeton University" might not be dominant. Thus, evaluating the magnitude of such accomplishments seems difficult and subjective.

Instead, we propose an approach to recognize the most descriptive Wikipedia categories based on categorical coherence. We observe that the figures in a precise category presumably are mutually similar: the presidents of the United States presumably more closely resemble each other than American writers or Grammy Award recipients. Such categorical coherence could be evaluated via similarities in texts or Wikipedia links of corresponding members in the category and we believe an interesting category requires coherence to make it interesting. We rank descriptive level of Wikipedia categories according to their coherence, which yield an overall agreement of 88.27\% compared with human wisdom.

Specifically, we make following contributions in this paper:

\begin{itemize}
\item
We create vector representations for people with Wikipedia pages using LDA topic modeling and Deepwalk. These representations position each person as a point in a high-dimensional space, facilitating similarity comparisons. Deepwalk representations trained on Wikipedia links are proved more valuable in our experiment than LDA topic modeling which is based on the text content.

\item
We present a new approach to identifying the most salient categories associated with
Wikipedia entities, based on the use of vector representations. We evaluate using different distance measures and coherence criteria to show what is best at quantifying descriptive level of Wikipedia categories. 

\item
Through human wisdom collected from Crowdflower, we verify that our notion of category appropriateness generally jibes with that of human reviewers. Indeed, we have fashioned a game app testing how often users agree with our algorithmically-chosen categorization.

\end{itemize}

The rest of this paper proceeds as follows: In Section 2 we review related work. Section 3 demonstrates the procedure of collecting human reviews. In Section 4 we describe Chinese menu choices in constructing our models. In Section 5 we show experiment results and corresponding analysis.

\section{Related Work}
Entity ranking gains popularity since better rankings boost the performances of search engines, resulting in faster and more precise information retrieval. Wikipedia seems to be a good playground. The problem of ranking web pages could be easily reduced to Wikipedia entity ranking, plus that Wikipedia has a large collection of entities of different types \cite{zaragoza2007ranking} and Wikipedia contains valuable texts, human annotated tags, enriched links and a great structure to analyze ranking effectiveness. Certain ranking can serve as a pivot for extensibility or analysis \cite{kaptein2010entity,lewandowski2011ranking}  or be used to answer queries in named entity recognition \cite{vercoustre2008entity}. Additionally, retrieving real-life knowledge of reputations, fames and historical significance from entity ranking is also valuable \cite{skiena2013s}.

Traditional ranking algorithms on Wikipedia basically consider two parts. One part focuses on information provided by raw text, including length of pages, word occurrences and topic distributions. LDA is among the most valuable approaches in such tasks \cite{blei2003latent}. Topics from LDA highly agree with real tags when finding most important feature words of a page \cite{krestel2009latent}. The other part of ranking criteria relies heavily on links. Representatives include PageRank \cite{brin1998anatomy} and HITS \cite{kleinberg1999authoritative}. PageRank is a link analysis algorithm that assigns high numerical weighting to pages that are referred to by many other pages and the structure of weight distribution conclude the importance of web pages. HITS defines hubs to be pages that have links pointing to many authority pages, serving as another important criteria in ranking. Recent work of Deepwalk \cite{Perozzi:2014:DOL:2623330.2623732} uses truncated random walks to learn latent representations by encoding social relations in a continuous vector space, which can be easily exploited by statistical models.

On the other hand, human annotated tags for pages benefits many related tasks. On Wikipedia, performance of entity ranking is improved by utilizing Wikipedia categories \cite{vercoustre2008using}. Links and topic difficulty prediction together with category information greatly boost the performance of entity ranking \cite{pehcevski2010entity}. Additionally, Wikipedia categories can be used to boost search results in an ad hoc way \cite{kaptein2009using}. Researchers also found that it is possible to analyze consistent semantic relationships in the tags \cite{chernov2006extracting} and corresponding latent representations of raw text would help tag recommendations \cite{krestel2009latent}. Given these facts, we believe a reversed application to rank Wikipedia categories based on latent representations of pages would help summarize the content in the text thus provides better and more precise descriptions of the Wikipedia pages.

\section{Collecting human wisdom}
Our experiment mainly focus on Wikipedia pages of historical figures. It is clear that not all categories are created with equal descriptive power, most of them are correct but provide only trivial information. For instance, categories like ``1961 births" and ``Living people" record certain status of facts of a person but are usually meaningless; categories like ``Harvard Law School alumni" provide better idea of a person but usually that is not enough to summarize a single aspect of one’s life. We here define descriptive power of a category to be the ability to construct analogous connections between people within same categories. The easier we can distinguish a person as well as other people with a single category tag, the more descriptive power we have on that category. We expect to find categories like ``Presidents of the United States" to be listed at the top of our ranking of descriptive power as these categories most accurately captures one’s historical/cultural significance.

We start a project on Crowdflower, a leading people-powered data enrichment platform, to collect human reviews of the most descriptive Wikipedia categories on the top-500 most famous people according to ``ranking of significance” described in \cite{skiena2013s}.

\begin{figure*}[!htb]
\centering
\includegraphics[width=1.0\textwidth]{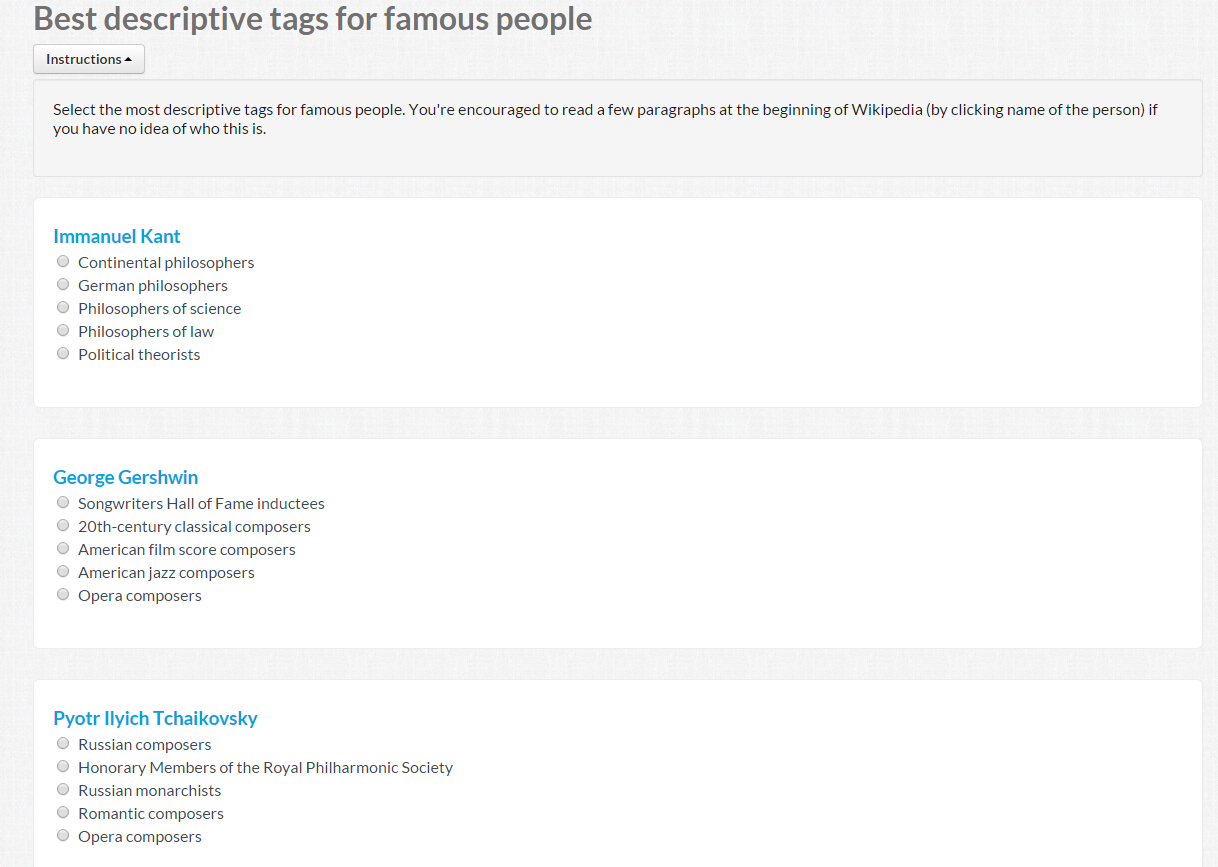}
\caption{Sample question to collect human wisdom on Crowdflower. Volunteers are required to pick the most important and descriptive one among these 5 choices.} 
\label{cf}
\end{figure*}

For each Wikipedia person we manually select 4 categories with good descriptive power plus 1 random category to make up a question with 5 choices. Example of questions are shown in Figure \ref{cf}.

Each answer makes a clarification that one choice is dominating the other four. If we collect multiple answers for a single question, categories with higher descriptive power will have more votes, thus the distribution indicate overall descriptive power of each Wikipedia category. For each question we collect answers from 20 volunteers. In total we gathered 10,000 answers from 176 volunteers, covering 1076 distinct Wikipedia categories. 

As another part of data collection, we extracted raw texts from all valid English Wikipedia pages and created adjacent lists for them. To avoid unexpected information from Wikipedia categories, we carefully removed all links / texts that appear in the reference of a main Wikipedia page. This procedure helps us collect texts / links of 4,517,721 pages. We also extracted corresponding Wikipedia categories for future uses.

\section{Chinese menu of ranking model}
In this section we will describe how we calculate categorical coherence. The procedure involves 4 major steps: Generating feature vectors using latent representations from text or links; Measuring distances in vector space; Definition of close neighbors; Calculating categorical coherence based on observations of close neighbors. We will have multiple choices for each step in the experiment and we expect to find the best combinations to solve this task.

\subsection{Generating feature vectors}
We focus on converting text / links of Wikipedia pages into vector representations. We experimented on 2 methods in our tests.

\subsubsection {LDA model}
\begin{figure}[!htb]
\centering
\includegraphics[width=0.5\textwidth]{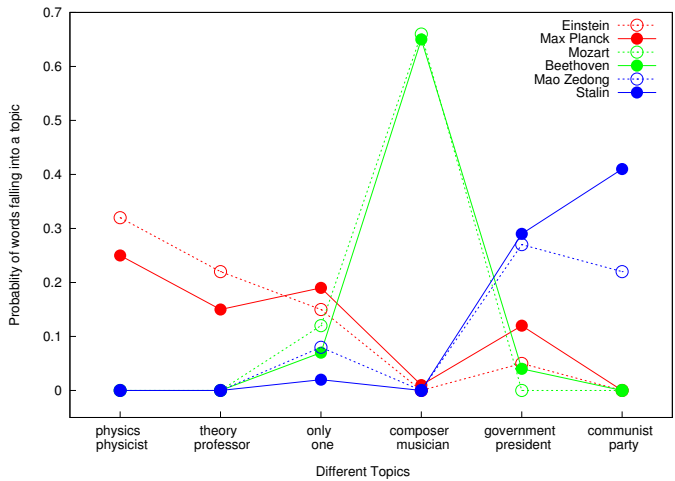}
\caption{Examples of entities and top six related topics in LDA method. We display top two representative words in each topic. The distribution of topics for Wikipedia people differs a lot for party leaders, musician and physicists. However, historical figures with comparable contributions or related professional fields are much more similar than the others.}
\label{cartoon2}
\end{figure}

LDA model provides probability distribution of possible topics. It is based on co-occurrence of words and has an advantage that the output of each topic would be easy for human beings to read and understand. We train LDA model for the all available pages in our data collection with 5000 topics, converting each Wikipedia page in the corpus into a probability distribution of possible topics. Figure \ref{cartoon2} shows an example of top related topics for some historical figures. Pages have higher probability to fall into same topics should be considered more similar thus close in high-dimensional vector space. 

\subsubsection {Deepwalk model}
Deepwalk is an online algorithm that creates statistical representation of graph by learning via random walks in the graph. Walks are considered as sentences metaphor and generate latent dimensions according to adjacent list. With a hierarchical Softmax layer these latent dimensions will be finally converted into vector representations. In our experiment, we propose that Wikipedia pages sharing more common links will sit closer in Deepwalk embedding space since random walks in corresponding pages visit through very similar paths. Groups with large fraction of links between corresponding Wikipedia people will indicate stable relationships on similarity as random walks have lower chances to step out of the group. We use the package described in \cite{Perozzi:2014:DOL:2623330.2623732} with 128 output dimensions to train Deepwalk embedding on adjacent lists of all Wikipedia pages. 

\subsection{Measuring distances}
Feature vectors are usually considered as points in high-dimensional space. Thus, distance between vectors may following the definition of Manhattan distance (L1 normalization) or Euclidean distance (L1 normalization). Plus, Cosine similarity are frequently used to measure similarities between feature vectors.

Additionally, since LDA represents probability distributions, we include Kullback–Leibler divergence and Jensen–Shannon divergence in our distance metrics.

\subsection{Definition of close neighbors}
Distance between feature vectors indicate similarity between corresponding Wikipedia people. However, such representations are not linear – pairs with twice the distance do not necessarily mean half the similarity. On the other hand, observation shows that only pairs within a certain range show strong signals of similarity. Such ranges are not pre-defined and usually it is correlated with the density of feature vectors in a certain area of the embedding space.

We here propose two approaches to define close neighbors. One is to limit the count of close neighbors, considering only the closest $K$ neighbors in vector space to be ``close" neighbors. Since relationship of close neighbors is not always reversible under this definition, this strategy will usually create asymmetric results. The other is to pick close neighbors by distance, marking all neighbors within a certain distance of $D$ as close. With this strategy, if a point in space is semi-isolated then nothing would be considered similar to it. Both two approaches are reasonable and will be considered as a hyper-parameter in our experiment.

\subsection{Ranking criteria}
We propose two methods to quantify how descriptive a Wikipedia category is.

\subsubsection{Conductance}
The ``conductance" of a category is defined as the ratio of close neighbors inside the category and those outside the category. This is a simple and direct measurement. Category with more close neighbors inside seems to be more stable since close neighbors are sharing similarities with ``members inside this category". However, this method does not consider the size of the category well. For larger categories with more corresponding people, it is even harder to keep all close neighbors inside the category / to guarantee category members are close, thus reducing the ``conductance" of the category.

Let $C_{cat}$ be the conductance of a category, then we have:
\[
C_{cat} = \displaystyle\sum_{\substack{X \in cat \\ Y \in cat \\ (X, Y) \in closeneighbors}} / \displaystyle\sum_{\substack{X \in cat \\ (X, Y) \in closeneighbors}}
\]

\begin{figure}[!htb]
\centering
\includegraphics[width=0.5\textwidth]{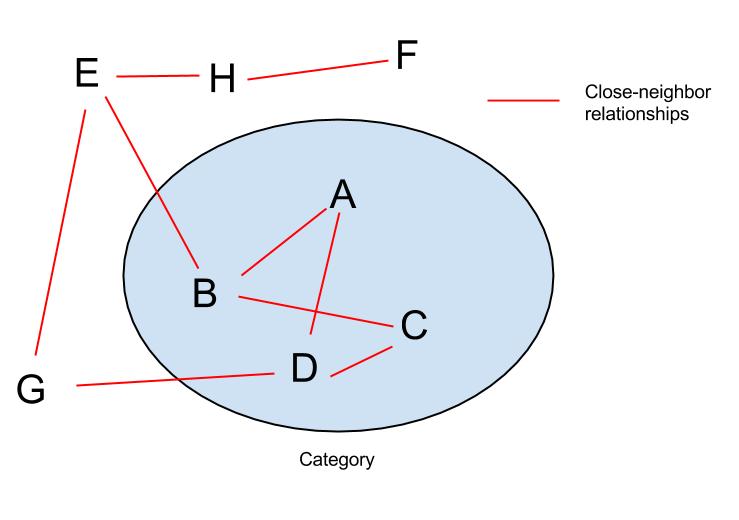}
\caption{Examples of calculating conductance. The category has 4 entities and 4 close-neighbor relationships inside. However, by the definition of ``close" there are 2 entities (E, G) which are close to one of the inside members with a total of 3 close-neighbor relationships that are not inside the category. The conductance of this category is $\frac{4}{4+3} \approx 0.5714$.}
\label{conductance}
\end{figure}

Figure \ref{conductance} shows an example of calculating conductance. 

\subsubsection{Surprise level}
The other measurement is named ``surprise level". This metric focuses on balancing the effect of both size of the category and the probability of inside-category pairs become close neighbors. We assume that the probability of ``a pair is inside a category" is independent from the probability of ``a pair shares close-neighbor relationship", then for an entity $A$ in the category the probability of its close neighbor $B$ belongs to the category will be:
\[
P_{cat} = \frac{\displaystyle\sum_{X \in cat}}{\displaystyle\sum_{X}}
\]

If we randomly pick $C_A$ close neighbors of entity $A$, the probability of having $K$ close neighbors in the group will be:

\[
P(cat, K, C_A) = \binom{C_A}{K} ( {P_{cat}}^K * (1 – P_{cat}) ^ {C_A – K} )
\]

We then use the real number of close neighbors $C_A$ and close neighbors that are in the category $G_A$ from observations in vector space and the surprise level of the category from observer $A$ will be defined as:

\[
S_{cat, A} = \displaystyle\sum_{X \ge G_A} P(cat, X, C_A)
\]

We then average the surprise level for all observers in the category to get the categorical surprise level. Example in Figure \ref{conductance} will result in following values:

\[
P_{cat} = \frac{4}{8} = 0.5
\]
\[
S_{cat, A} = S_{cat, C} = \binom{2}{2} P_{cat} = 0.25 \\
\]
\[
S_{cat, B} = S_{cat, D} = \binom{3}{2} P_{cat} + \binom{3}{3} P_{cat}  = 0.5 \\
\]
\[
S_{cat} = \frac{0.25 + 0.5 + 0.25 + 0.5}{4} = 0.375 \\
\]

\subsection{Chinese menu combination}
In our experiment, we will conduct a grid search with following choices:
\begin{itemize}
\item Feature vectors: (LDA, Deepwalk)
\item Distance function: (L1, L2, Cosine, KL (for LDA) and JS (for LDA))
\item Close neighbors: (Count and Distance). For count strategy we will test using parameters of 5, 10, 25, 50 and 100. For distance strategy we sort pair-wise distances and choose thresholds to keep an average of 5, 10, 25, 50 and 100 close neighbors correspondingly. 
\item Measurement: (Conductance and Surprise level)
\end{itemize}

\section{Evaluation results and analysis}

We will rank all Wikipedia categories in our data collection using our Chinese menu combination model. However, final quality of our ranking will be evaluated according to the agreement with human annotations, i.e. the ranks of 1067 categories that appear in at least 1 question in human annotations. 

\subsection{Basic evaluation}
A single answer in our collection votes for the most descriptive categories among 5 possible choices, thus each answer actually indicates 4 comparisons between the voted one and the remaining 4. If the voted answer ranks top in our ranking compared with other 4 choices, then our ranking get 1 point for making 4 correct judges; if the voted answer ranks second then we miss one comparison and score 0.75 point ans so on. For each answer, we will gain (5 – i) * 0.25 point towards the final score where i is the relative rank among 5 choices in our ranking. Rough overall accuracy will be calculated as:

\[
Acc_{rough} =  \displaystyle\frac{\sum_{i} Score\ for\ answer_i}{|Answers|}
\]

Where the total number of answers is 10,000 (500 questions * 20 answers each question).

\subsection{Maximum possible accuracy?}
Human reviews are not perfect and there are possible conflicts between answers. Observations on the data collection show that some categories are highly correlative and confusing as they often co-appear and they share similar descriptive power. Table \ref{coappear} lists top 10 most confusing category pairs.

\begin{table}[!htb]
\centering
\small
\begin{tabular}{| l | r |}
\hline
\ Category pairs & Co-Prob \\
\hline
\ French Open champions & \\
\ Wimbledon champions & 100.00\% \\
\hline
\ Australian film actors & \\
\ Australian television actors & 100.00\% \\
\hline
\ Association football forwards & \\
\ Brazilian footballers & 86.60\% \\
\hline
\ Holocaust perpetrators & \\
\ Nazi Germany ministers & 86.60\% \\
\hline
\ National Basketball Association All-Stars & \\
\ Parade High School All-Americans (boys' basketball) & 81.65\% \\
\hline
\ Eastern Orthodox saints & \\
\ People celebrated in the Lutheran liturgical calendar & 75.59\% \\
\hline
\ American novelists & \\
\ American short story writers & 67.94\% \\
\hline
\ American jazz singers & \\
\ Traditional pop music singers & 67.61\% \\
\hline
\ American rhythm and blues singer-songwriters & \\
\ American soul singers & 67.36\% \\
\hline
\ African-American rappers & \\
\ Pseudonymous rappers & 62.90\% \\
\hline
\end{tabular}
\caption{10 most confusing category pairs in our questions. \emph{Co-Prob} of two categories $A$ and $B$ is defined as the geometric mean of $P(A|B)$ and $P(B|A)$. Since candidate choices are manually picked from existing Wikipedia categories, such co-occurrence reflect some preference in background knowledge. }
\label{coappear}
\end{table}

Since the existence of confusing category pairs, human reviews cannot perfectly reach the maximum possible accuracy of 100\% and the descriptive power will reflect in the distribution of voted answers. Here we propose an improved overall accuracy which consider the imperfectness of human reviews. We construct a graph where nodes are Wikipedia categories and directed edges show that people votes more to one answer if both appear in the same question. We run a topological sort algorithm to figure out that the best possible ``cheating" score is 8456.25, which means based on the gold standard from Crowdflower human reviews, about 15.5\% answers will never be correct.

The improved overall accuracy is now set to:
\[
Acc_{improved} = \displaystyle\frac{\sum_{i} Score\ for\ answer_i}{|Best Cheating Score|}
\]

\subsection{Influence of feature vectors}
We first discuss the influence of choosing different feature vectors in Table \ref{features}.

\begin{table}[!htb]
\centering
\begin{tabular}{| c | c  c | c | c |}
\hline
\ & & & \multicolumn{2}{c|}{Feature vector} \\
\cline{4-5}
\ & & & LDA & Deepwalk \\
\hline
\ \multirow{14}{*}{Best} & \multicolumn{2}{c|}{Overall Accuracy} & 86.70\% & \textbf{88.27\%} \\
\cline{2-5}
\ & \multirow{4}{*}{Parameters} & Distance & L1 & L2 \\
\ & & Closeness & Count & Count \\
\ & & Avg Neighbors & 25 & 50 \\
\ & & Ranking & SL & SL \\
\cline{2-5}
\ & \multirow{5}{*}{Agreement} & 1st & 44.07\% & 46.47\% \\
\ & & 2nd & 25.10\% & 24.79\% \\
\ & & 3rd & 16.29\% & 14.97\% \\
\ & & 4th & 9.11\% & 8.32\% \\
\ & & 5th & 5.44\% & 5.44\% \\
\cline{2-5}
\ & \multirow{5}{*}{Distance} & 5 & 0.1352 & 1.0092 \\
\ & & 10 & 0.1972 & 1.0704 \\
\ & & 25 & 0.3091 & 1.1740 \\
\ & & 50 & 0.4223 & 1.2823 \\
\ & & 100 & 0.5639 & 1.4366 \\
\hline
\ \multirow{7}{*}{Average} & \multicolumn{2}{c|}{Overall Accuracy} & 75.67\% & \textbf{78.01\%} \\
\cline{2-5}
\ & \multicolumn{2}{c|}{Std Deviation} & 0.062 & 0.044 \\
\cline{2-5}
\ & \multirow{5}{*}{Agreement} & 1st & 33.55\% & 36.67\% \\
\ & & 2nd & 23.86\% & 22.81\% \\
\ & & 3rd & 18.13\% & 17.51\% \\
\ & & 4th & 13.93\% & 13.71\% \\
\ & & 5th & 10.53\% & 9.30\% \\
\hline
\end{tabular}
\caption{Performance of each feature vector. \emph{Best} rows describe the best achieved performance, including improved overall accuracy, model parameters (distance measurement, close neighbor definition, average count close neighbors and ranking criteria), distribution of points gaining and corresponding distance distribution in vector spaces. \emph{Average} rows provide information of average accuracy, standard deviation and distribution of average points gaining.}
\label{features}
\end{table}

Deepwalk basically outperforms LDA. Deepwalk actually making 1.5\% more answers correct and points gained from agreement with the 1st vote from human reviews are also higher. Distance distribution both vectors are stable since the gap between each threshold (5, 10, 25, 50, 100) is increasing reasonably, indicating that points in high-dimensional spaces are not too dense. Deepwalk also win considering average performances among all Chinese menus parameters, with higher accuracy and lower standard deviation.

\subsection{Influence of distance measurement}
Table \ref{distance} shows statistics of how different distance measurements change the performance.

\begin{table}[!htb]
\centering
\begin{tabular}{| c | c  c | c c c c c |}
\hline
\ & & & \multicolumn{5}{c|}{Distance measurements} \\
\cline{4-8}
\ & & & L1 & L2 & Cos & KL & JS \\
\hline
\ \multirow{6}{*}{Best} & \multicolumn{2}{c|}{Accuracy} & 87.94\% & \textbf{88.27\%} & 86.70\% & 85.73\% & 86.45\% \\
\cline{2-8}
\ & \multirow{5}{*}{Agree} & 1st & 46.09\% & 46.47\% & 44.07\% & 43.71\% & 43.99\% \\
\ & & 2nd & 24.91\% & 24.79\% & 25.10\% & 24.73\% & 24.12\% \\
\ & & 3rd & 14.91\% & 14.97\% & 16.29\% & 14.25\% & 17.33\% \\
\ & & 4th & 8.57\% & 8.32\% & 9.11\% & 12.45\% & 9.41\% \\
\ & & 5th & 5.52\% & 5.44\% & 5.14\% & 4.87\% & 5.14\% \\
\cline{2-8}
\hline
\ \multirow{7}{*}{Average} & \multicolumn{2}{c|}{Accuracy} & 75.41\% & 77.94\% & \textbf{78.87\%} & 75.98\% & 76.10\%\\
\cline{2-8}
\ & \multicolumn{2}{c|}{Std Deviation} & 0.062 & 0.044 & 0.033 & 0.063 & 0.063\\
\cline{2-8}
\ & \multirow{5}{*}{Agree} & 1st & 33.51\% & 36.53\% & 37.07\% & 34.03\% & 33.96\%\\
\ & & 2nd & 23.53\% & 22.93\% & 22.78\% & 23.50\% & 23.85\%\\
\ & & 3rd & 18.18\% & 17.48\% & 18.38\% & 18.30\% & 18.19\%\\
\ & & 4th & 14.06\% & 13.77\% & 13.37\% & 13.77\% & 13.63\% \\
\ & & 5th & 10.71\% & 9.29\% & 8.40\% & 10.39\% & 10.36\% \\
\hline
\end{tabular}
\caption{Performance of each distance measurement. JS and KL are uniquely used in LDA model. \emph{Best} rows describe the best achieved performance, including improved overall accuracy, distribution of points gaining. \emph{Average} rows provide information of average accuracy, standard deviation and distribution of average points gaining.}
\label{distance}
\end{table}

From our observations there are no big gaps discovered between L1 normalization and L2 normalization for best / average performances. Cosine similarity is working well on average cases but it does not win to produce the best performances. KL divergence and JS divergence, though specifically designed for probability distributions, do not yield better performances in our tasks. This phenomena probably indicates the redundancy in feature vectors so that no matter which normalization function was chosen, there will be a factor that preserve the property of making similar points close enough.

\subsection{Influence of defining close neighbors}
\begin{figure}[!htb]
\centering
\includegraphics[width=0.5\textwidth]{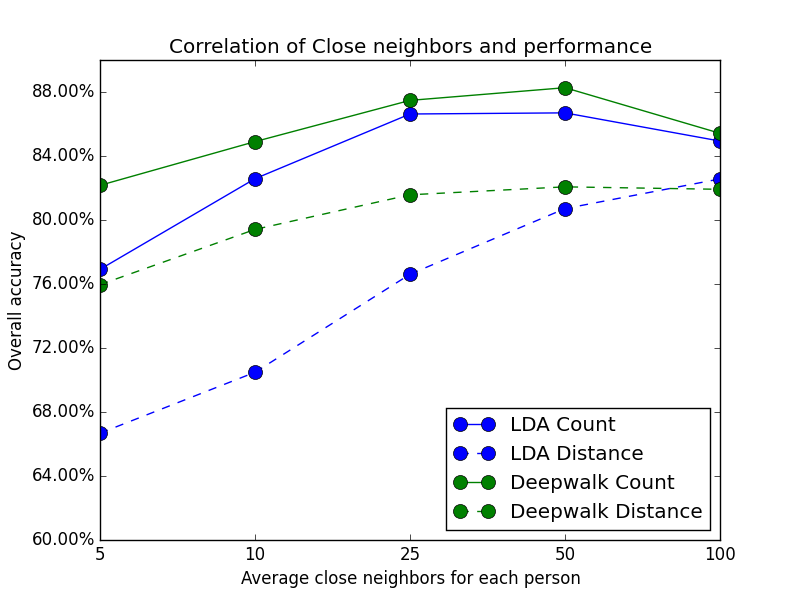}
\caption{Different definition of close neighbors and corresponding best performance. We experimented strategies of making a fixed number of close neighbors for each point as well as creating a comparable number of close neighbors overall using a distance limitation. Fixed number strategy outperforms the distance definition.}
\label{close}
\end{figure}

Figure \ref{close} shows 10 different definitions of close neighbors, with corresponding performances, including 5 using count as threshold and 5 using distance as threshold

It is clear that judging close neighbors using only the value of distance is worse. The reason is that density of local surrounding of semi-isolated points (e.g. person with few introduction text and Wikipedia links) is much lower than those frequently mentioned historical figures. Setting threshold to be a certain diameter will create uneven distribution of close neighbors, thus lower quality and stability of similarity measurement.

On the other hand, both count and distance threshold shows a peak within the range of 25 to 50 average close neighbors, which is approximately 0.005\% to 0.01\% of the whole data collection (Total number of people’s pages on Wikipedia about 557,596). Neither increasing nor decreasing this value will give better performances. We believe such criteria would still work even under more sophisticated circumstance, e.g. considering most famous 50,000 people since points in the embedding space are usually evenly distributed.

\subsection{Influence of ranking criteria}
Table \ref{measure} lists comparisons between ``Conductance" and ``Surprise level". One interesting observation is, the best performance of all conductance measurement happens when creating an average of 100 close neighbors for each point in feature vector space. However, we discovered that Surprise level measurement still outperforms Conductance even with much fewer close neighbors, indicating that the size of category should be carefully examined during the procedure. 

\begin{table}[!htb]
\centering
\begin{tabular}{| c | c c | c | c |}
\hline
\ & & & \multicolumn{2}{c|}{Ranking criteria} \\
\cline{4-5}
\ & & & Conductance & Surprise level \\
\hline
\ \multirow{6}{*}{Best} & \multicolumn{2}{c|}{Overall Accuracy} & 81.08\% & \textbf{88.27\%} \\
\cline{2-5}
\ & \multirow{5}{*}{Agreement} & 1st & 36.49\% & 46.47\% \\
\ & & 2nd & 26.23\% & 24.79\% \\
\ & & 3rd & 19.24\% & 14.97\% \\
\ & & 4th & 11.11\% & 8.32\% \\
\ & & 5th & 6.93\% & 5.44\% \\
\cline{2-5}
\hline
\ \multirow{7}{*}{Average} & \multicolumn{2}{c|}{Overall Accuracy} & 77.63\% & \textbf{80.34\%} \\
\cline{2-5}
\ & \multicolumn{2}{c|}{Std Deviation} & 0.059 & 0.038 \\
\cline{2-5}
\ & \multirow{5}{*}{Agreement} & 1st & 34.51\% & 37.77\% \\
\ & & 2nd & 24.85\% & 23.20\% \\
\ & & 3rd & 18.49\% & 18.65\% \\
\ & & 4th & 13.03\% & 13.82\% \\
\ & & 5th & 9.11\% & 6.57\% \\
\hline
\end{tabular}
\caption{Performance of each ranking criteria. \emph{Best} rows describe the best achieved performance, including improved overall accuracy, model parameters (distance measurement, close neighbor definition, average count close neighbors and ranking criteria), distribution of points gaining and corresponding distance distribution in vector spaces. \emph{Average} rows provide information of average accuracy, standard deviation and distribution of average points gaining.}
\label{measure}
\end{table}

\subsection{Error analysis}
We try to make a deep analysis based on the best ranking, which is generated using Deepwalk embedding, L2 normalization, each person having 25 close neighbors and measured via Surprise level. We list all categories with more votes but less surprise level in our data collection. Table \ref{cate} shows the top 10 of them:
\begin{table}[!htb]
\centering
\begin{tabular}{| c | r | r |}
\hline
\ Category & Count & Probability\\
\hline
\ Prime Ministers of the United Kingdom & 162 & 85.26\% \\
\ Presidents of the United States & 116 & 82.86\% \\
\ American pop singer-songwriters & 111 & 58.42\% \\
\ The Beatles members & 80 & 72.73\% \\
\ American rhythm and blues singers & 75 & 62.50\% \\
\ American male professional wrestlers & 54 & 60.00\% \\
\ First Ladies of the United States & 52 & 86.67\% \\
\ American rock singers & 50 & 83.33\% \\
\ American rock guitarists & 50 & 71.43\% \\
\ American horror writers & 12 & 62.50\% \\
\hline
\end{tabular}
\caption{Categories disagree most with votes. \emph{Count} shows number of times human vote for this category among all answers containing this choice. \emph{Probability} shows the chance of this category being answered whenever it appears.}
\label{cate}
\end{table}

As we can see, 3 out of 10 categories (Prime Ministers of the United Kingdom, Presidents of the United States, First Ladies of the United States) are political leaders whose titles are so recognizable that they bestowed enough to distinguish this person from the others. However, such categories have a long history – it is quite possible that there are less similarity between U.S. presidents in 1800s and U.S. presidents after 2000 except the title itself and Deepwalk did not find supporting evidences from Wikipedia links. ``The Beatles Members" plays a special role since the size of the category is too small while the descriptive power is unbelievably large. The remaining categories are rough and generalized, sometimes with confusions, which makes it hard to process.

\subsection{Sample Results}
Here we listed top 100 Wikipedia categories discovered by our algorithms in Table \ref{samp} using best Deepwalk model with 50 close neighbors, L2 normalization and Surprise level (represented in logarithm as S-Level). 

\begin{table*}[!htb]
\centering
\begin{tabular}{| l | l | l | l |}
\hline
\ Category & S-Level &  Category & S-Level \\
\hline
\ World Aquatics Championships medalists in swimming & -280.98 & PGA Tour golfers & 	-274.52 \\
\ Australian rugby league players & 	-269.04 & Chess grandmasters & -256.92 \\
\ Teachtaí Dála & -250.48 & Medalists at the World Figure Skating Championships &	-238.43 \\
\ German footballers & 	-228.79  & India One Day International cricketers & 	-226.46 \\
\ American astronauts & 	-224.48 & Italian footballers & -221.90 \\
\hline
\ NASCAR drivers & 	-221.48 & Australia Test cricketers & 	-212.77 \\
\ Filipino television actors &	-210.48 & Spanish footballers &	-206.72 \\
\ England Test cricketers &	-203.06 & South Korean television actors &	-200.92 \\
\ Members of the Knesset & -198.02 & Olympic medalists in athletics (track and field) & 	-191.42 \\
\ Members of the Australian House of Representatives & 	-190.78 & Malayali actors &	-189.00 \\
\hline
\ Japanese voice actors  &	-185.33 & United States men's international soccer players &	-183.62 \\
\ Prime Ministers of Japan &	-183.23 & All-Australians (AFL) &	-178.09 \\
\ Olympic medalists in gymnastics  &	-177.27 & Blues Hall of Fame inductees &	-177.22 \\
\ World Series of Poker bracelet winners &	-173.10 & Recipients of the Knight's Cross &	-172.49 \\
\ Leaders of the Communist Party of China &	-170.48 & French painters 	& -168.76 \\
\ Dutch footballers & 	-167.80 & Mexican footballers &	-166.27 \\
\hline
\ Brazilian footballers & -165.41 & Italian painters & -164.61 \\
\ International Tennis Hall of Fame inductees &	-163.84 & American mixed martial artists &	-161.55 \\
\ UK MPs 2005–2010 &	-161.44 & National Basketball Association All-Stars  &	-160.44 \\
\ English footballers &	-156.94 & Jamaican reggae musicians &	-155.56 \\
\ Portuguese monarchs &	-154.01 & Italian popes &	-153.84 \\
\hline
\ Presidents of Mexico & 	-153.20 & France international footballers &	-151.68 \\
\ Imperial Roman consuls &	-149.97 & Filmfare Awards winners &	-149.49 \\
\ African-American rappers &	-148.59 & Australian soccer players &	-148.16 \\
\ American country singers 	& -148.08 & Explorers of Antarctica &	-147.57 \\
\ Scottish footballers &	-146.34 & Ancient Greek philosophers & 	-143.96 \\
\hline
\ National Hockey League All-Stars &	-142.87 & World Boxing Council Champions 	& -141.58 \\
\ MotoGP riders 	& -141.36 & Professional Darts Corporation players &	-141.35 \\
\ Russian composers 	& -140.85 & American mobsters of Italian descent &	-140.83 \\
\ Members of the Queen's Privy Council for Canada &	-140.20 & 24 Hours of Le Mans drivers &	-138.47 \\
\ United States Supreme Court justices  &	-137.06 & Apostles of The Church of Jesus Christ of Latter-day Saints &	-134.75 \\
\hline
\ Pakistani politicians & -134.66 & Worldcon Guests of Honor &	-133.57 \\
\ Analytic philosophers &	-132.82 &  Pakistan Test cricketers &	-132.72 \\
\ House of Holstein-Gottorp-Romanov & -132.00 & Burials at Roskilde Cathedral &	-134.10 \\
\ National Baseball Hall of Fame inductees &	-131.86 & Tennis players at the 2012 Summer Olympics &	-131.19 \\
\ New Zealand international rugby union players & 	-130.72 & Union Army generals &	-130.16 \\
\ Japanese pop singers & 	-130.00 & Prime Ministers of New Zealand & -129.00 \\
\hline
\ American television chefs &	-127.00 & Saints from the Holy Land &	-126.24 \\
\ Rurik Dynasty &	-125.14 & Japanese professional wrestlers &	-124.65 \\
\ Members of the United States House of Representatives from California  &	-124.30 &  American female pornographic film actors &	-123.35 \\
\ Portugal international footballers & -123.33 & 20th-century mathematicians & 	-122.45 \\
\ Will Eisner Award Hall of Fame inductees & -121.78 & Presidents of Argentina & 	-121.45  \\
\hline
\ Hungarian monarchs & -121.41 & College men's basketball head coaches in the United States &	-120.00 \\
\ Nobel laureates in Economics &	-119.82 & American professional wrestlers &	-119.69 \\
\ Japanese emperors & 	-118.64 & England international rugby union players &	-117.98 \\
\ Burials at Riddarholmen Church &	-117.55 & Continental philosophers &	-116.88 \\
\ Confederate States Army generals &	-116.08 & Members of the Bundestag 	& -115.88 \\
\hline
\ TVB veteran actors & -115.87 & Candidates for President of the Philippines & -115.85 \\
\ American printmakers &	-115.36 & Members of the Scottish Parliament 1999–2003 &	-114.84 \\
\ English snooker players &	-114.83 & Italian film directors &	-114.29 \\
\ Prime Ministers of France & 	-114.21 & American architects &	 -113.28 \\
\ Members of the House of Representatives of the Netherlands &	-112.84 & Polish monarchs & 	-111.55 \\
\hline
\end{tabular}
\caption{Top 100 most descriptive categories detected by our model. Experiment is conducted on 557,596 Wikipedia pages of historical figures. High surprise level shows strong similarities between people in the same category. Some categories discovered by the algorithm have low semantic indications in the text  but encoded historical / cultural importance (e.g. Burials at Roskilde Cathedral).}
\label{samp}
\end{table*}

\section{Conclusion}
In our work we proposed models to rank Wikipedia categories according to their descriptive power. We experimented two models to convert texts of links in Wikipedia to feature vectors and tried different definition of closeness that reflects similarities between entities. We then calculate descriptive power based on categorical coherence, which reflects how well a category keeps its inside members close from latent representations of feature vectors. Our models naturally extend to analyzing pages in different languages, and also to extend to other classes of entities like locations (i.e. cities and countries) and organizations (companies and universities) and we are able to identify similar individuals for suggesting friends in social networks, or even matching algorithms pairing up roommates or those seeking romantic partners.

We collected human reviews indicating descriptive power Wikipedia categories from Crowdflower. We tested our models on approximately 600,000 historical figures pages from Wikipedia. Deepwalk model trained using Wikipedia links yielded the best overall accuracy of 88.27\% in our evaluation, showing a good recovery of importance of Wikipedia categories.

The final target of our application is not only focusing on identifying similarities on Wikipedia. Considering possible applications of finding similar people via their personal webpage or resume (which focus on text) or their social network friends list (using graph structures), we are glad to see that our models can be applied to various types of text and graphs. Such utility could help improve algorithms of online recommending systems. 

We provide sorted list of Wikipedia categories that best summarize the historical/cultural significance of people, which can be used to better explain similarity between historical figures or serve as a pivot to improve search results. We collect knowledge to understand the importance of these particular strong or overrepresented features in our analysis. 

Future work includes parameterizing our methods so we can capture different tradeoffs between different models as topic-based analysis and links analysis focus on different aspects of the pages. 

\bibliography{myrefs}
\bibliographystyle{IEEEtran}

\end{document}